\documentclass[11pt,a4paper]{article}
\usepackage[hyperref]{acl2019}
\usepackage{times}
\usepackage{latexsym}
\usepackage{graphicx}
\usepackage{graphics}
\usepackage{url}
\usepackage{amsmath}
\usepackage{url}
\usepackage{enumitem}
\usepackage{color}
\usepackage{booktabs}
\usepackage{multirow}
\usepackage{amsmath}
\usepackage[T1]{fontenc}
\usepackage{natbib}
\usepackage{paralist}

\aclfinalcopy 

\title{Learning to Plan and Realize Separately \\for Open-Ended Dialogue Systems}

\author{Sashank Santhanam\textsuperscript{\rm 1}, Zhuo Cheng\textsuperscript{\rm 1}, Brodie Mather\textsuperscript{\rm 2}, Bonnie Dorr\textsuperscript{\rm 2}, \\ \textbf{Archna Bhatia\textsuperscript{\rm 2}, Bryanna Hebenstreit\textsuperscript{\rm 2}, Alan Zemel\textsuperscript{\rm 3},} \\ \textbf{Adam Dalton\textsuperscript{\rm 2}, Tomek Strzalkowski\textsuperscript{\rm 4} and Samira Shaikh\textsuperscript{\rm 1}}  \\
University of North Carolina, Charlotte, NC, USA\textsuperscript{\rm 1}\\
Institute for Human and Machine Cognition (IHMC), Ocala, FL, USA\textsuperscript{\rm 2}\\
State University of New York, Albany, NY, USA\textsuperscript{\rm 3} \\
Rensselaer Polytechnic Institute, Troy, NY, USA\textsuperscript{\rm 4} \\
\texttt{\{ssantha1, samirashaikh\}@uncc.edu}\textsuperscript{\rm 1}}

\date{}

\begin{document}
\maketitle
\begin{abstract}
Achieving true human-like ability to conduct a conversation remains an elusive goal for open-ended dialogue systems. We posit this is because extant approaches towards natural language generation (NLG) are typically construed as end-to-end architectures that do not adequately model human generation processes.   
To investigate, we decouple generation into two separate phases: planning and realization. In the planning phase, we train two planners to generate plans for response utterances. The realization phase uses response plans to produce an appropriate response. Through rigorous evaluations, both automated and human, we demonstrate that decoupling the process into planning and realization performs better than an end-to-end approach.
\end{abstract}

\section{Introduction}
\label{intro}

Recent advancements in the area of generative modeling have helped increase the fluency of generative models. However, several issues persist: coherence of output and the semblance of mere repetition/hallucination of tokens from the training data \cite{moryossef2019step, wiseman2017challenges}. 
One reason could be that the generation task is typically construed as an end-to-end system. This is in contrast to traditional approaches, which incorporate a sequence of steps in the NLG system, including content determination, sentence planning, and surface realization \cite{reiter1994has, reiter2000building}. A review of literature from psycholinguistics and cognitive science also provides strong empirical evidence that the human language production process is not a monolith \cite{dell1985positive,bock1996language,bock2007persistent,kennison2018psychology}.

\begin{figure}[t]
    \centering
    \includegraphics[width=\columnwidth, height=6cm]{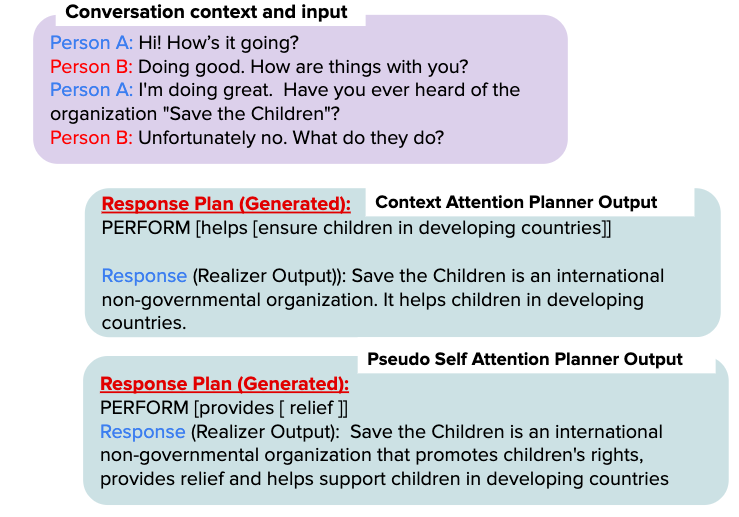}
    \caption{Example conversation between two speakers A \& B where the response for the speaker B is generated based on the response plan from two learned planners: Context Attention and Pseudo Self Attention.}
    \label{fig:example}
\end{figure}

Prior approaches have indeed incorporated content planning into the NLG system, for example data-to-text generation problems \cite{puduppully2019data, moryossef2019step} as well as classic works that include planning, based on speech acts \cite{cohen1979elements} (for an in-depth review c.f. \cite{garoufi2014planning}). 
Our work closely follows these prior approaches, with one crucial difference: our planners are not based on dialogue acts or speech acts. 

Consider the example in Fig.~\ref{fig:example}. An input utterance by Person B, a statement (\emph{Unfortunately no.}), followed by a question (\emph{What do they do?}), can be effectively responded to using plans, learned and generated, prior to the realization phase. The realization output can then include the mention of \emph{provides relief}, consistent with the generated plan (\emph{PERFORM [provides [relief]]}). 

Dialogue acts \cite{stolcke2000dialogue} (e.g., statements, questions), by their nature, encompass a wide variety of realized output, and hence cannot sufficiently constrain the language model during the generation process. Research has addressed this issue by adapting existing taxonomies \cite{stolcke2000dialogue} towards their own goals \cite{wu2018towards,oraby2017may}. We instead use an adapted and extended form of lexical-conceptual structures (LCSs) to help constrain the realization output more effectively \cite{Dorr:1994}.

Our work makes the following contributions: \\
\begin{inparaenum}
    \item[\textbullet] We investigate the impact of separating planning and realization in open-domain dialogue and find that the approach produces better responses per automated metrics and detailed human evaluations.\\
    \item[\textbullet]  We propose the use of LCS-inspired representations based on asks and framings, which in turn are grounded in conversation analysis literature, to generate plans, instead of using dialogue acts.    \\
    \item[\textbullet]  We release corpora annotated with plans for all utterances, using three planners, including symbolic planners and attention-based planners. 
\end{inparaenum}

\section{Related Work}

\textit{\textbf{Open-Ended Dialogue Systems:}}
 
Transformer models \cite{vaswani2017attention} and large transformer-based language models such as GPT, GPT-2, XLNet, BERT \cite{radford2018improving,radford2019language,DBLP:conf/nips/YangDYCSL19,devlin2019bert} have helped achieve the SOTA performance across several natural language tasks. However, these models do not achieve the same level of consistent performance on generative modeling tasks as opposed to language understanding tasks \cite{ziegler2019encoder,edunov2019pre}. 
Wolf \emph{et al.} \citeyearpar{wolf2019transfertransfo} propose a transfer learning approach that fine tunes large pretrained language models and achieves SOTA scores on the PERSONA-chat dataset \cite{golovanov2019large} and in the CONVAI2 competition \cite{Dinan:2019,yusu2018nips}.  Keskar \emph{et al.} \citeyearpar{keskar2019ctrl} introduce a large-scale conditional transformer model that improves generation based on control codes. 

Our training paradigm is consistent with existing research that constrains large-scale language models across generation tasks \cite{rashkin2019towards,DBLP:conf/emnlp/UrbanekFKJHDRKS19} and yields controllable text generation \cite{shen-etal-2019-select,zhou2017end}, with one key difference: we learn to plan and realize separately. Accordingly, we overview planning based approaches next. 

\textit{\textbf{Planning-Based Approaches:}}
A standard component of traditional NLG systems is a planner \cite{reiter2000building}. Prior work leverages intent and meaning representations (MR) to understand the content of the message \cite{young2013pomdp}, but largely in task-oriented as opposed to open-ended dialogue systems \cite{he2018decoupling}. Novikova \emph{et al.} \citeyearpar{novikova2017e2e} propose the E2E challenge and use MRs to show lexical richness and syntactic variation. Similarly, Gardent \emph{et al.}  \citeyearpar{gardent2017creating} focus on structured data (e.g. DBpedia) to generate text in the WebNLG framework. Moryossef \emph{et al.} \citeyearpar{moryossef2019step} use an explicit symbolic component for planning in a neural data to text generation system that allows controllable generation. Along with conversational intents, dialogue acts are also used for natural language understanding (NLU) in task-oriented systems \cite{li2019end, peskov-etal-2019-multi}. 

In contrast to these prior approaches, our work uses more in-depth meaning representations for open-domain dialogue systems based on lexical conceptual structures (explained in Section 3.1).

\section{Approach}

\subsection{NLU using Asks and Framing}

The representation we use to generate plans leverages \emph{asks} and \emph{framings} based on conversation analysis literature
\cite{PomerantzFehr2011,Sacks1992,Schegloff2007}. An \textit{ask} is closely related to the notion of a request \cite{Zemel2017Texts}.   
Perhaps most importantly, an ask elicits relevant responses from the recipient. \textit{Framing} refers to linguistic and social resources used to persuade the recipient of an ask to comply and perform the requested social action. Put another way, an ask creates a social obligation to respond, while framing provides an adequate basis for compliance with the ask.   

\begin{figure}[h]
  \centering
  \small
    \includegraphics[width=1.0\columnwidth]{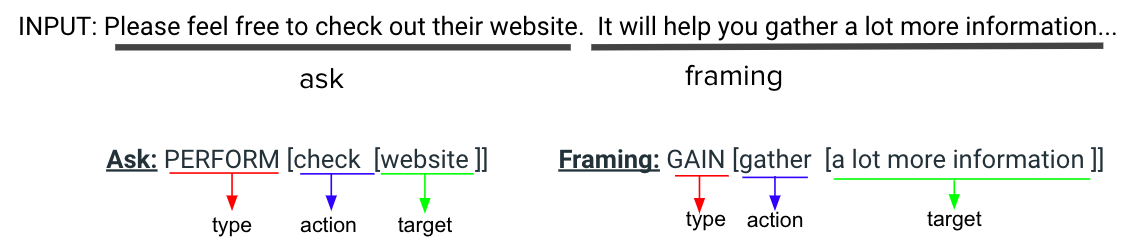}
  \caption{Example of ask and framing representations used as training for generation of Response Plans.}
  \label{fig:ask_example}
\end{figure}
\begin{figure*}[t]
  \centering
  \small
    \includegraphics[width=1.0\textwidth]{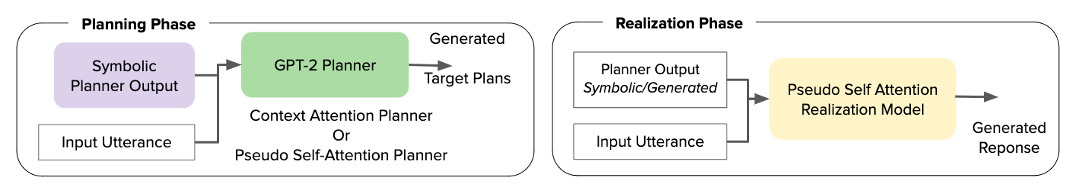}
  \caption{Architecture diagram of our system consisting of two phases: Planning and Realization. The Planning phase (Context and Pseudo Self Attention) encodes the input sequence and symbolic planner input to produce the response plans. The Realization phase uses the response plan and input utterance to generate the response}
  \label{fig:arch}
\end{figure*}

In Fig.~\ref{fig:ask_example}, we show the ask/framing representational formalism that serves as the basis of our response plans. Here the \emph{ask} is a request to PERFORM the action of \emph{check out the website}. The perceived risk or reward (or \emph{framing}) for this request is that, upon performing the action, one may GAIN something, i.e., \emph{gather a lot more information}. We use two types of \emph{asks}: GIVE (provide something or information) and PERFORM (perform an action), and two types of \emph{framings}: GAIN (gain some benefit) and LOSE (lose benefit or resource). This preliminary ontology was motivated by conversation analysis literature \cite{sacks1978simplest,curl2008contingency,epperson2008reports}: by treating utterances as actions, we are able to establish what each utterance seeks to accomplish and how a sender motivates the recipient in terms of the benefits and costs of compliant responses.

\subsection{Method}

Our goal is to generate an informative response to the input utterance by first generating an appropriate \textbf{Response Plan}. We train two components separately (c.f. Fig.~\ref{fig:arch}). In the \textit{\textbf{Planning Phase}}, we experiment with generating plans in three ways: \\
\textit{1. Symbolic Planner}: Foremost, we need to extract plans automatically from utterances. To accomplish this goal, our symbolic planner adapts lexical representations previously used for language analysis \cite{dorr2020detecting} to the problem of constructing \textbf{Response Plans}. 
We use lexical conceptual structures and basic language processing tools \cite{Gardner2017AllenNLP,manning-EtAl:2014:P14-5} for parsing the input, identifying the main \textbf{action}, identifying the arguments (or \textbf{targets}), and applying semantic-role labeling.  Fig.~\ref{fig:ask_example} presents ask/framing examples (type, action and target).

Once response plans are identified for all utterances in a given corpus using the symbolic planner, we need to address \textit{automated generation} of such plans. Using the asks and framings as annotated data for a ``silver'' standard,\footnote{Dorr \emph{et al.} \citeyearpar{dorr2020detecting} report precision of 69.2\% in detecting asks/framings.} we train models to learn to generate ``Response Plans'' that are encoded with the same representation format used for asks/framings. We use the language modeling paradigm and use a large pre-trained model (GPT-2) \cite{radford2019language} with the transformer architecture and the self-attention mechanism \cite{vaswani2017attention}. We fine-tune this language model with the constraint of the input utterance and the plan for this input utterance, and train it to produce the plan for the response utterance. We adopt the fine-tuning approach specified by Ziegler \emph{et al.} \citeyearpar{ziegler2019encoder} and train two specific models (CTX and PSA) described below.  

\textit{2. Context Attention Planner (CTX):} based on the encoder/decoder architecture. In this model, the decoder weights are initialized with the pre-trained weights of the language model. However, a new context attention layer is added in the decoder that concatenates the conditioning information to the pre-trained weight. The conditioning information, in our case, is the plan for the input utterance. 
        
\textit{3. Pseudo Self Attention (PSA):} Proposed by Ziegler \emph{et al.} \citeyearpar{ziegler2019encoder}, PSA injects conditioning information from the encoder directly into the pre-trained self attention (similar to the ``zero-shot'' model proposed by Radford \emph{et al.} \citeyearpar{radford2019language}).

In the \textbf{\textit{Realization Phase}}, we generate responses by utilizing the response plan generated from the planning phase as well as the input utterance. We expect a more guided generation of responses that are constrained by the response plan. In this phase, we only experiment with the Pseudo Self attention (PSA) model, based on  Ziegler \emph{et al.} \citeyearpar{ziegler2019encoder}, who demonstrate that PSA outperforms other approaches on text generation tasks.  We use nucleus sampling to overcome some of the drawbacks of beam search \cite{Holtzman2020The}.

\subsection{Corpora}

Our choice of corpora is driven by the presence of information elicitation and persuasive strategies in the utterances (i.e., asks and framings). 

Accordingly, we experiment with the AntiScam \cite{li2019end} and Persuasion for Social Good \cite{DBLP:conf/acl/WangSKOYZY19} corpora. \textbf{AntiScam} contains dialogues about a customer service scenario and is specifically crowdsourced to understand human elicitation strategies. \textbf{Persuasion for Social Good} corpus contains interactions between workers who are assigned the roles of persuader and persuadee, where the persuader attempts to convince the persuadee to donate to a charity. 

All utterances in these corpora are first annotated through the Symbolic Planner (c.f. Section 3.2) to gauge suitability based on the presence of asks and framings. In Table~\ref{stats}, we provide descriptive statistics of the corpora; we find an adequate number of ask/framing types (GIVE, PERFORM, GAIN, LOSE). In cases where there are no asks/framings or the symbolic planner fails to detect them, we use the default action RESPOND.

\begin{table}[t]
\small
\centering
\begin{tabular}{lcc}
\toprule

                     & AntiScam & \begin{tabular}[c]{@{}c@{}}PSG\end{tabular} \\ \toprule
Number of Dialogues     & 220      & 1017                                                                  \\ \midrule
\begin{tabular}[l]{@{}l@{}} Avg. Conversation Length \end{tabular}     & 12.45    & 10.43                                                                 \\ \midrule
\begin{tabular}[l]{@{}l@{}} Avg. Utterance Length \end{tabular} & 11.13    & 19.36                                                                 \\ \midrule
Number of GIVE          &       2192   &       11587                                                                \\ \midrule
Number of PERFORM       &      1681    &    7335                                                                    \\ \midrule
Number of GAIN          &       70   &          399                                                             \\\midrule 
Number of LOSE          &      73    &      588     \\ \midrule     

Number of RESPOND          &      4376    &  8078         \\ \bottomrule     
\end{tabular}
\caption{Statistics of AntiScam and Persuasion for Social Good (PSG), with annotated asks and framings. Avg. conversation length - average number of turns in each conversation; Avg. utterance length - average length of a turn in a conversation}
\label{stats}
\end{table}

\subsection{Implementation}
We implement the models using Open-NMT \cite{opennmt} and the PyTorch framework.\footnote{\url{https://pytorch.org/}} We use publicly available GPT-2 model \cite{radford2019language} with 117M parameters, 12 layers and 12 heads in our implementations. The input utterances and the plans are tokenized using byte-pair encoding to reduce vocabulary size \cite{sennrich2015neural}. Both phases are trained separately. In the Planning Phase, the \emph{plan for the input} utterance along with the input utterance is used to generate the \emph{response plan} for the response utterance; in the Realization Phase, the response plan and input utterance are input to the model to generate the response. In both planning and realization phase, separation tokens are added (e.g. $<$plan$>$), as is common practice for transformer inputs \cite{devlin2019bert,wolf2019transfertransfo}.  We use Adam optimizer \cite{kingma2014method} with a learning rate of $0.0005$ and $\beta_1=0.9$ and $\beta_2=0.98$. During decoding, we use nucleus sampling both in the planning and realization phase. 
All models are trained on two TitanV GPU and take roughly 15 hours each to train the planner and realization component. The trained models and the codebase are available at \url{https://github.com/sashank06/planning_generation}

\section{Evaluation of Approach}

The results reported in these subsections were obtained by combining both corpora and dividing randomly in a ratio of 80/10/10 for the training, testing, and validation set. 

\begin{table*}[t]
\centering
\small
\begin{tabular}{llllllll}
\toprule
Model             & BLEU-1          & BLEU-2          & BLEU-3          & BLEU-4          & CIDER           & ROUGE@L         & METEOR          \\ \toprule \toprule
Context Attention (CTX) & 0.1097          & 0.0714          & 0.0571          & 0.0506          & 0.5053          & 0.1677          & 0.3444         \\ \midrule
Pseudo-Self Attention (PSA)         & \textbf{0.1342} & \textbf{0.0886} & \textbf{0.0672**} & \textbf{0.0578**} & \textbf{0.6506} & \textbf{0.2108} & \textbf{0.3447} \\ \bottomrule
\end{tabular}
\caption{Automated Metrics on performance of models in the Planning Phase. ** indicates $p<0.01$}
\label{plannerautomated}
\end{table*}

\subsection{Planning Phase Evaluation}
 
This evaluation focuses on investigating the efficacy of the two automated planners (Context Attention (CTX) and Pseudo-Self Attention (PSA)) in learning to generate response plans. 
\subsubsection{Automated Metrics}
\emph{Are the automated planners able to faithfully learn how to generate the response utterance plans?} To investigate, we compare the performance of the CTX and the PSA planner with the symbolic planner output (which is our silver standard reference) using common automated metrics  
Table \ref{plannerautomated}: BLEU \cite{papineni2002bleu}, METEOR \cite{banerjee2005meteor}, ROUGE \cite{lin2004rouge}, CIDEr \cite{vedantam2015cider} on the test set. We use the library by Sharma \emph{et al.} \citeyearpar{sharma2017nlgeval}. We find that PSA was able to achieve higher word overlap metrics with respect to the silver standard. We conducted an in-depth analysis of the CTX and PSA planner output on the entire testing set. We found that the PSA model was more likely to produce ask actions that matched the ground truth, resulting in higher scores on the automated metrics. 

\subsubsection{Human Evaluation}

\begin{table}[h]
\centering
\small
\begin{tabular}{lllll}
\toprule
   & \begin{tabular}[c]{@{}l@{}}CTX \end{tabular} & \begin{tabular}[c]{@{}l@{}}PSA \end{tabular}& \multicolumn{1}{l}{Both} & \multicolumn{1}{l}{Neither} \\ \toprule \toprule
Q1 & 38.75\%                               & 26.25\%                                   & 25\%                     & 10\%                        \\ \midrule
Q2 & 27.5\%                                & 20\%                                      & 23.75\%                  & 28.75\%                     \\ \midrule
Q3 & 22.5\%                                & 17.5\%                                    & 41.25\%                  & 18.75\%                     \\ \midrule
Q4 & 32.5\%                                & 31.25\%                                   & 10\%                     & 26.25\%   \\ \bottomrule                 
\end{tabular}
\caption{Human Evaluation results on the performance of the planner component. \textbf{Q1:} Which model plan is better suited for generating a response?; \textbf{Q2:} Which model has the more appropriate ask/framing type?; \textbf{Q3:} Which model has the more appropriate ask/framing action with respect to the type?; \textbf{Q4:} Which model has the more informative ask/framing target?}
\label{planner-human}
\end{table}
Evaluation using automated metrics provides limited evidence for the ability to automatically generate plans; we do not know if these plans are actually useful in a realization task. The question then is: \emph{How well-suited are the automatically learned plans for the task of generating responses?} 

\textbf{Study 1:} We asked two experts in linguistics to independently rate 40 randomly sampled plans from the test set. For context, we provided the input utterance and its plan produced by the symbolic planner. Their task was to choose which of the learned response plans was better suited to the realization task (CTX, PSA, Both or Neither). They also evaluated the plan constituents: (\textbf{type}, \textbf{action} and \textbf{target}). We randomized the presentation order of the planner outputs across questions to avoid ordering/learning effects \cite{medin1994presentation}. We find an inter-rater agreement \cite{shrout1979intraclass} of 0.5 ($p<0.001$) between the linguists.

Table \ref{planner-human} shows the results from Study 1. From \textbf{Q1}, we find that CTX planner is better suited to generate an appropriate response over the PSA planner. Similarly, through \textbf{Q2}, \textbf{Q3}, and \textbf{Q4}, we find that the CTX planner is better able to generate the appropriate ask/framing types, actions, and targets. We also find that the linguists rated Neither plan was suited to generate a response 10\% of the time. Put differently; the automatically generated plans would work 90\% of the time to generate an appropriate utterance in the realization phase. The learned plans have trouble associating an appropriate ask/framing type and target (28.75\% and 26.75\%) but perform better with the ask/framing action (18.75\% Neither rating). 

This evaluation compares the automatic planners against one another, but \emph{how well do the planners compare to the silver standard (symbolic planner)?}

\begin{table}[]
\small
\centering
\begin{tabular}{lllll}
\toprule
                         & CTX  & PSA  & \begin{tabular}[c]{@{}l@{}}Symbolic\\ Planner\end{tabular} & Both \\ \toprule \toprule
\multirow{2}{*}{Quality} & 30\% & X    & 35\%                                                       & 35\% \\ \cmidrule{2-5}
                         & X    & 35\% & 22\%                                                       & 43\% \\ \bottomrule
\end{tabular}
\caption{Human evaluation results comparing CTX and PSA planner  separately to the Symbolic Planner}
\label{planner_human_eval_comparison}
\end{table}

\textbf{Study 2:} We asked the same linguistic experts to independently determine which amongst two plans (symbolic vs. each automated planner) would be more appropriate to generate a response. This study design is consistent with prior studies in dialogue evaluation \cite{mei2017coherent, serban-etal-2016-generating}.  Table \ref{planner_human_eval_comparison} presents the results from Study 2. 
 
We find that experts prefer the plans produced by the symbolic planner over the CTX output but not over the PSA planner output. Inter-annotator agreement \cite{shrout1979intraclass} between the experts for this study was 0.54. While Study 1 compared CTX and PSA planner outputs against one another, Study 2 compared CTX and PSA outputs against the silver standard. As we observe from the automated metrics (Table \ref{plannerautomated}), PSA model plans are more faithful to the ground truth, e.g., higher BLEU 1-4 scores than CTX model plans. Since PSA planner outputs are more faithful to the ground truth, this may be why human judges rate them as preferable more often when compared against ground truth.

\textit{\textbf{Planning Phase Evaluation Findings:}} To summarize this evaluation section, we find: PSA  outperforms the CTX planner on automated metrics. This finding is consistent with the results from Ziegler \emph{et al.} \citeyearpar{ziegler2019encoder}. From Study 1, we find that both the planners are able to generate appropriate plans, with the appropriate ask/framing type, action, and target for the realization phase, a large proportion of the time. From Study 2, we find that when compared to the silver standard plans, PSA planner output is preferred over the CTX planner. 

\begin{table*}[t]
\centering
\small
\begin{tabular}{llllll}
\toprule
Realizer Input                                                                                   & Dataset                                                              & BLEU    & Diversity &  Length  & BERT-score\\ \midrule \midrule
\multirow{2}{*}{No Plan}                                                             & AntiScam                                                             & 0.0658  & \textbf{0.0067}          &         7.168   & 0.841    \\ \cmidrule{2-6}
                                                                                        & \begin{tabular}[c]{@{}l@{}}PSG\end{tabular} & 0.1149  &    \textbf{0.0049}       &            13.713  &  0.845   \\ \midrule
\multirow{2}{*}{Symbolic Planner}                                                       & AntiScam                                                             & \textbf{0.1814}  &     0.0062      &             6.245  & \textbf{0.844}   \\ \cmidrule{2-6}
                                                                                        & \begin{tabular}[c]{@{}l@{}}PSG\end{tabular} & \textbf{0.1992}  &   0.0038        &            11.982     & \textbf{0.848}  \\ \midrule
\multirow{2}{*}{\begin{tabular}[c]{@{}l@{}}Context Attention\\ Planner\end{tabular}}    & AntiScam                                                             & 0.0705 &    0.0064       &            7.298  & 0.84      \\ \cmidrule{2-6}
                                                                                        & \begin{tabular}[c]{@{}l@{}}PSG\end{tabular} & 0.1027  &   0.0043        &            14.088  &   0.847  \\ \midrule
\multirow{2}{*}{\begin{tabular}[c]{@{}l@{}}Pseudo Self \\ Attention Planner\end{tabular}} & AntiScam                                                             & 0.0692  &      0.0065     &              \textbf{ 7.553} & 0.838    \\ \cmidrule{2-6}
                                                                                        & \begin{tabular}[c]{@{}l@{}}PSG\end{tabular} & 0.1253  &   0.0045        &            \textbf{15.128} & 0.847\\ \bottomrule       
\end{tabular}
\caption{Automated metric results on the responses generated on the test set of both corpora. }
\label{dialog-automated}
\end{table*}

\subsection{Realization Phase Evaluation}
While the previous section focuses on evaluating the ability to generate plans automatically, we do not yet know \emph{whether separating the generation process into planning and realization produces better responses than an end-to-end system?}

Thus, we compare four approaches towards realizing a response given an input utterance (through the Pseudo-Self Attention fine-tuned realization algorithm): (1) \textbf{No Planner} model which receives input utterance but no plan as input; (2) \textbf{Symbolic Planner based Generation}: This model receives the plan from symbolic planner output; (3) \textbf{CTX Planner-Based Generation}: This model receives the CTX plan; (4) \textbf{PSA Planner-Based Generation}: This model receives the PSA plan. 

\subsubsection{Automated Metrics}
\label{dialog_automate}
Prior research has shown that most automated metrics have little to no correlation to human ratings on NLG tasks \cite{dialogue-eval,santhanam-shaikh-2019-towards}; however, they may provide some standard of reference to evaluate performance. We report the following metrics: 
\begin{inparaenum}[(i)]
    \item BLEU \cite{papineni2002bleu}
    \item length of responses, with the understanding that models that are able to generate longer responses are better 
    \item following, Mei \emph{et al} \citeyearpar{mei2017coherent}, we report the diversity metric \cite{diversity2016}. Diversity is calculated as the number of distinct unigrams in the generation scaled by the total number of generated tokens \cite{mei2017coherent,li2016persona}.
    \item BERT-Score \cite{bert-score} metric, an embedding-based score which has shown greater correlation to human ratings.
\end{inparaenum}

Table \ref{dialog-automated} reports on the automated evaluation against the ground truth utterances. We find that on both corpora and across all metrics except Diversity, incorporating plans as an additional input to the realization phase helps achieve a higher score than having No Planner. From Table \ref{dialog-automated}, we find that the realizer without any plans is able to achieve higher diversity, but the difference is not statistically significant.

\subsubsection{Human Evaluation}
\label{dhumaneval}
Since automated metrics are not the most informative indicators of quality of generated responses, thorough human evaluation is necessary. We \emph{investigate if humans prefer the responses generated by the planner-based models over those generated without the plan (No Planner)}.
We conducted two human evaluation studies by recruiting workers from Amazon Mechanical Turk service with strict quality control criteria: workers should have at least 90\% HIT approval rate and at least 1000 approved HITs. In each survey, workers are asked to evaluate responses on these metrics, following Novikova \emph{et al.} \citeyearpar{novikova2018rankme}:
\begin{inparaenum}[(i)]
    \item \textit{Appropriateness:} determines whether response aligns with the topic of the conversation and the input utterance.
    \item \textit{Quality:} determines the overall quality in terms of grammatical correctness, fluency, and adequacy
    \item \textit{Usefulness:} determines if the response is highly informative to generate a response.
\end{inparaenum}

\begin{table}[h]
\centering
\small
\begin{tabular}{llll}
\toprule
                                                         Realizer Input  & \begin{tabular}[c]{@{}l@{}}Appropri-\\ ateness\end{tabular} & Quality       & \begin{tabular}[c]{@{}l@{}}Useful-\\ ness\end{tabular} \\ \toprule \toprule
No Plan                                                 & 2.54                                                       & 2.61          & 2.58                                                   \\ \midrule
\begin{tabular}[c]{@{}l@{}}Symbolic Planner \end{tabular} & 2.51                                                       & 2.5           & 2.53                                                   \\ \midrule
\begin{tabular}[c]{@{}l@{}}CTX Planner\end{tabular} & \textbf{2.34}                                              & \textbf{2.38} & \textbf{2.38}                                          \\ \midrule
\begin{tabular}[c]{@{}l@{}}PSA Planner\end{tabular}      & 2.59                                                       & 2.5           & 2.51 \\ \bottomrule                                                 
\end{tabular}
\caption{Average ranking of realized output from four different planners, lower score is better}
\label{humaneval-ranking}
\end{table}

\begin{figure*}[t]
    \includegraphics[width=\textwidth, height=4.7cm]{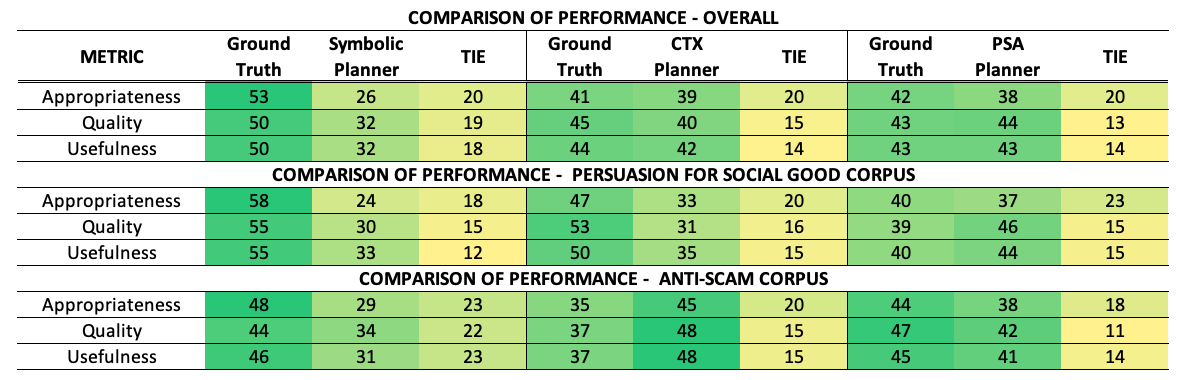}
    \caption{Comparison of ground truth reference with realized output from each model that receives learned plans as input: Symbolic, CTX or PSA. Higher values (shown as \%)/darker color represent better performance.} 
    \label{comparison-groundtruth}
\end{figure*}

\textit{\textbf{Study 1:}} We tasked 30 crowd-sourced workers to rank order the four model responses from best to worst. We randomly sampled 60 examples from the test set with an even 50\% split (30 examples each) between the Persuasion for Social Good and AntiScam corpora. We chose the best to worst ranking mechanism since it has shown greater consistency and agreement amongst workers on tasks related to dialogue evaluation over other evaluation designs (e.g. Likert scales) \cite{10.1145/3313831.3376318, kiritchenko-mohammad-2017-best}. The presentation order of model outputs for each question was again randomized to avoid learning effects \cite{medin1994presentation}. Table \ref{humaneval-ranking} 
demonstrates the average rank position (1$=$Best, 4$=$Worst) obtained by each model. We find using the plans generated by the CTX planner helps generate better responses. On the metrics of quality and usefulness, we find that incorporating planning as additional input performs better than no plan (i.e. end-to-end system). 

\textit{\textbf{Study 2:}} In this study, we evaluate \emph{how well the generated responses compare to the ground truth.} The ground truth references are those produced by humans in the PSG and Anti-Scam corpora. We recruited 11 MTurk workers with the same crowdsourcing quality controls as Study 1. For the same randomly sampled 60 examples from Study 1, workers were asked if they prefer the ground-truth response, the response generated from the three planners, or both, on the three chosen metrics. This study design is also consistent with prior work \cite{mei2017coherent}. Workers were blinded to the source of the response (ground truth or generated) and were presented the responses in a randomized order across all questions to avoid ordering effects.

Fig.~\ref{comparison-groundtruth} shows the results (higher value/darker color is better): we find that responses generated from the symbolic planner as input do not perform well when compared to the ground truth. In other words, the proportion of time that the ground truth response is preferred over that generated by the symbolic planner is significant (e.g. 53\% vs. 26\% on the Appropriateness metric overall). 

We find that on all three metrics, the responses generated using CTX and PSA plans help generate responses that are comparable to the responses produced by humans (ground truth). We also find that the PSA planner-based responses perform better overall and on the Persuasion for Social Good corpus. Surprisingly, the CTX planner based responses performs better than Ground Truth utterances for the Anti-Scam corpus (45\%, 48\% and 48\% of the time preferable vs. ground truth response 35\%, 37\% and 37\% on the three metrics, Appropriateness, Quality, and Usefulness, resp.). We explain this unexpected finding in the next sub-section (Section 4.3). 

\textit{\textbf{Realization Phase Evaluation Findings:}} To summarize this evaluation subsection, we find that the Symbolic Planner-realized output outperforms the CTX, PSA, and No Planner output on the automated metrics of BLEU and BERT-score. Importantly, the CTX planner-realized output has a higher rank in terms of overall preference in human evaluation than the other models (c.f. Table~\ref{humaneval-ranking}). We also find that human-generated utterances (ground truth) are preferred overall (c.f Fig.~\ref{comparison-groundtruth}) than the model outputs. We found inter-rater consistency and agreement scores to be $>$0.6 on average across the metrics (full tables are reported in the Appendix).

\begin{figure}[t]
    \centering
    \vspace*{-.1in}
    \includegraphics[width=\columnwidth, height=6cm]{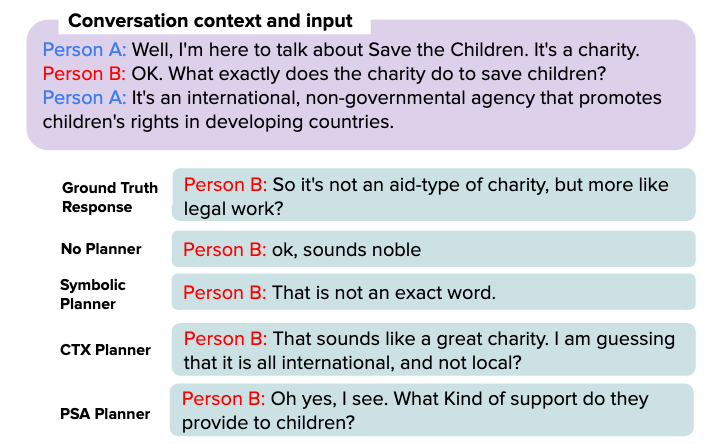}
    \caption{Sample outputs from realization phase with all variations of planner input, as well as the ground truth response from the corpus}
    \label{fig:sample_conversation}
\end{figure}

\begin{table*}[t]
\footnotesize
\centering
\begin{tabular}{@{}lll@{}}
\toprule
\multicolumn{1}{c}{\textbf{Issues Found}} & \multicolumn{1}{c}{\textbf{Input Utterance and Context}} & \multicolumn{1}{c}{\textbf{\begin{tabular}[c]{@{}c@{}}Generated Plan for Response/ \\ Generated Response\end{tabular}}} \\ \midrule
\multicolumn{3}{c}{Planning Phase} \\ \midrule
\midrule
\begin{tabular}[c]{@{}l@{}}\textit{Non-Informative} \\ \textit{Ask/Framing Target}\end{tabular} & \begin{tabular}[c]{@{}l@{}}The money goes directly to the organization in order to help. \\ The money will have a large impact. Would you like to have \\ the site info?\end{tabular} & PERFORM [see [ that ]] \\ \midrule
\begin{tabular}[c]{@{}l@{}}\textit{Wrong Ask/Framing } \\ \& \textit{type and action}\end{tabular} & \begin{tabular}[c]{@{}l@{}}no, that has never been my address.\end{tabular} & GIVE [give [ why got ]] \\ \midrule \midrule
\multicolumn{3}{c}{Realization Phase} \\ \midrule \midrule
\begin{tabular}[c]{@{}l@{}} \textit{Ignored} \\
\textit{Plan}
\end{tabular} & \begin{tabular}[c]{@{}l@{}}you certainly can, I am also able to take this payment\\ information as well.\end{tabular} & \begin{tabular}[c]{@{}l@{}}Plan: PERFORM [ find [your billing \\ date and names ]]\\ Response:  Okay, thanks!\end{tabular} \\ \midrule
 \begin{tabular}[c]{@{}l@{}} \textit{Grammatical} \\
\textit{Inconsistency}
\end{tabular} & \begin{tabular}[c]{@{}l@{}}Person A: The payment has not been processed. \\ Person B:  Oh why not?\end{tabular} & \begin{tabular}[c]{@{}l@{}}Plan: PERFORM[help[with that]]\\ Response: They help with that.\end{tabular} \\ \bottomrule
\end{tabular}
\caption{Summary table of issues found through a qualitative analysis of the planning and realization phase outputs}
\label{tab:issues}
\end{table*}

\subsection{Qualitative Analysis}
We conduct a qualitative evaluation of the outputs and present several cherry- and lemon-picked examples here. Additional examples of success and failure cases are uploaded in the Appendix.  
In the sample conversation shown in  Figure~\ref{fig:sample_conversation}, we find that realized outputs using CTX and PSA plans are more consistent with the context of conversation than the symbolic planner approach. Additionally, the 
No Planner output (an end-to-end system which does not get a plan as an additional input) produces an utterance that may not necessarily continue the conversation further. 

This example is also illustrative of the finding in Study 2 of the Planning Phase evaluation, where the crowdsourced workers rated the automated planner-based outputs better than the symbolic planner-based outputs (c.f. Fig.~\ref{tab:issues}). This might seem contradictory, as the CTX and PSA planners are trained on the silver standard data from the symbolic planner. We contend that this is due to the ability of automated planners (CTX and PSA) to generalize, an ability lacking in the symbolic planner. In such cases, as shown in Fig.~\ref{fig:sample_conversation}, the symbolic planner defaults to the RESPOND message plan, and this lead to generated output: \emph{That is not an exact word}, which is generic and off-topic. The symbolic planner could be improved to cover more cases; however, the effort would not be scalable.

While we find promising results for the automatically-generated planners in Sections 4.1 and 4.2, areas of improvement do exist (Table~\ref{tab:issues}):

\textit{\textbf{Non-Informative Ask/Framing Targets:}} We find several examples where the ask/framing targets are non-informative words (e.g. \emph{this, that}). Non-informative targets can cause the downstream realization process to generate an utterance that is, in turn, also non-informative. One example of such cases is shown in Row 1 of Table~\ref{tab:issues}. 

\textit{\textbf{Wrong Type and Action:}} Another planning phase issue category is that the constituents of plan representation (e.g., the ask/framing type and action) can be incorrect. As illustrated by the example in Table~\ref{tab:issues}, an ask target of \emph{why got} is incorrect.  Typically, we would expect to find a noun or a noun phrase as the ask/framing action (e.g., \emph{your billing date and names} as shown in the plan in Row 3).  

\textit{\textbf{Ignored Plan:}}
In the Realization phase, a typical issue is that the realizer may ignore the generated plan. As can be seen in Row 3 of Table~\ref{tab:issues}, the plan should constrain the response, and thus should contain phrases such as \emph{finding your billing date and names}. However, the generated response is instead a generic phrase \emph{Okay, thanks!}. 

\textit{\textbf{Grammatical inconsistencies:}}
We also note that there were cases where the grammar, e.g. pronoun usage, is inconsistent. For the example shown in Row 4 of Table~\ref{tab:issues}, we see that the generated response is \emph{They help with that.} whereas the conversation is between two persons; a generated response of \emph{I can help with that} would be more consistent with the context of the conversation. 

\section{Conclusion and Future Work}
We address the task of natural language generation in open-ended dialogue systems. We test our hypothesis that decoupling the generation process into planning and realization can achieve better performance than an end-to-end approach.

\textit{In the planning phase}, we explore three methods to generate response plans, including a Symbolic Planner and two learned planners, the Context Attention and Pseudo Self Attention models. Through linguist expert evaluation, we are able to determine the efficacy of the response plans towards realization. \textit{In the realization phase}, we use the Pseudo Self Attention model to make use of the learned response plans to generate responses.

\textit{\textbf{Our key finding through two separate human crowdsourced studies is that decoupling realization, and planning phases outperforms an end-to-end No Planner system across three metrics (Appropriateness, Quality, and Usefulness).}}

In this work, we have taken an initial step towards the goal of replicating human language generation processes. Thorough and rigorous evaluations are required to fully support our claims, e.g., by including additional metrics and more diverse corpora. In this work, we limit the types to  GIVE, GAIN, LOSE, and PERFORM. However, we do not restrict the ask action and target at all. Also, since our symbolic planner can be used to obtain silver standard training data, straightforward changes like adding additional lexicons would enable us to generalize to other corpora as well as include additional ask types in our pipeline. Another natural extension would be to explore training the planning and realization phases together in a hierarchical process \cite{fan2018hierarchical}. This would, in principle, further validate the efficacy of our approach.

 \section*{Acknowledgments}
This work was supported by DARPA through AFRL Contract FA8650-18-C-7881 and through Army Contract W31P4Q-17-C-0066. All statements of fact, opinion or conclusions contained herein are those of the authors and should not be construed as representing the official views or policies of DARPA, AFRL, Army, or the U.S. Government.

\bibliography{acl2019}
\bibliographystyle{acl_natbib}

\newpage
\appendix
\section{Supplementary Materials}
\subsection{Planner Output Analysis}
Table ~\ref{planner_performance} shows the performance of planners on the test set. We count the number of ask/framing types as well as the number of default response plan produced by each planner: GIVE, PERFORM, GAIN, LOSE and RESPOND from the testing set. 
\begin{table}[h]
\centering
\small
\begin{tabular}{llll}
\toprule
              & \begin{tabular}[c]{@{}l@{}}Symbolic\\ Planner\end{tabular} & \begin{tabular}[c]{@{}l@{}}CTX\\ Planner\end{tabular} & \begin{tabular}[c]{@{}l@{}}PSA\\ Planner\end{tabular} \\ \toprule \toprule
Num of GIVES   & 1248            & 1187        & 1146        \\ \midrule
Num of PERFORM & 815             & 1041        & 1129        \\ \midrule
Num of GAIN    & 44              & 35          & 29          \\ \midrule
Num of LOSE    & 66              & 35          & 49          \\ \midrule
Num of RESPOND & 969             & 842         & 789    \\ \bottomrule    
\end{tabular}

\caption{Distribution of different types of asks/framings in the test set of the planning component. Note: We found two asks produced by the CTX planner that ignored our ontology, which are excluded from our counts in this table}
\label{planner_performance}
\end{table}

\subsection{Inter-rater Consistency for Realization Phase}
We present the inter-rater consistency and agreement scores for the crowd-sourced worker studies we conducted during Realization Phase evaluation. The results presented were calculated using the \texttt{R} \textit{irr} package.\footnote{\url{https://cran.r-project.org/web/packages/irr/irr.pdf}}

\begin{table}[h]
\centering
\small
\begin{tabular}{lccc}
\toprule
            & Appropriateness & Quality & Usefulness \\ \toprule \toprule
Consistency & 0.42            & 0.65    & 0.67       \\ \midrule
Agreement   & 0.42            & 0.65    & 0.67      \\ \bottomrule
\end{tabular}
\caption{ICC-Consistency and Agreement Scores on the Appropriateness, Quality and Usefulness metrics obtained from Study 1 in the Realization Phase}
\label{ICC_study1_realization}
\end{table}

\begin{table}[h]
\centering
\small
\begin{tabular}{lcccc}
\toprule
            & Appropriateness & Quality & Usefulness \\ \toprule \toprule
Consistency 
& 0.65            & 0.60    & 0.58       \\ \midrule
Agreement  
& 0.62            & 0.59    & 0.55      \\ \bottomrule
\end{tabular}
\caption{ICC-Consistency and Agreement Scores on the Appropriateness, Quality and Usefulness metrics obtained from Study 2 in the Realization Phase}
\label{ICC_study2_realization}
\end{table}

\subsection{Additional Output Examples}

In this section, we give additional examples of conversations from our test set. Realization output based on each planner configuration of the system is included. In the Tables~\ref{tab:issues_supple}, we provide additional examples of the issues we found through manual inspection of the outputs.

\begin{figure}[h]
    \centering
    \includegraphics[width=\columnwidth, height=6cm]{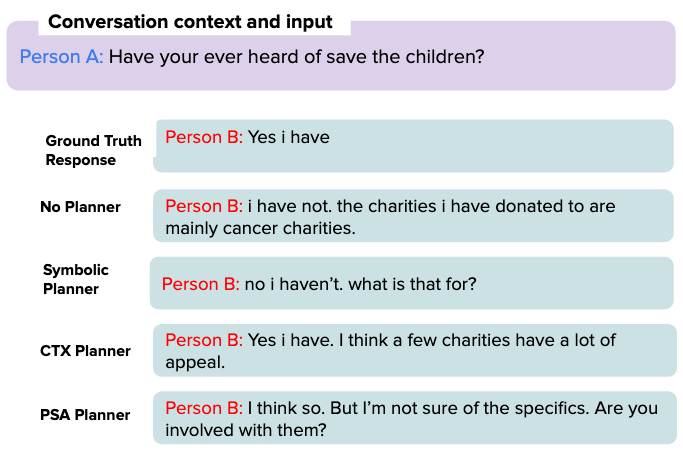}
    \caption{Example conversation between two speakers A \& B from test set. We can see that in the case of the realizer output from Symbolic Planner and PSA Planner, the responses also includes an ask (e.g. \emph{what is that for?} and \emph{Are you involved with them?} which may serve to carry the conversation further, as compared to other responses.}
    \label{fig:example_supple1}
\end{figure}

\begin{figure}[h]
    \centering
    \includegraphics[width=\columnwidth, height=6cm]{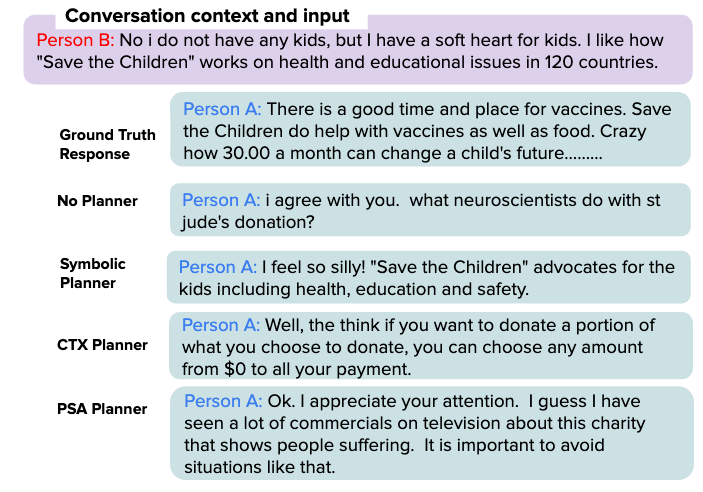}
    \caption{Another conversation between two speakers A \& B from our test set. The ground truth response in this case was lengthier than the typical response (consisting of 73 words) and has been shortened here for ease of presentation.}
    \label{fig:example_supple2}
\end{figure}

\begin{figure}[h]
    \centering
    \includegraphics[width=\columnwidth, height=7cm]{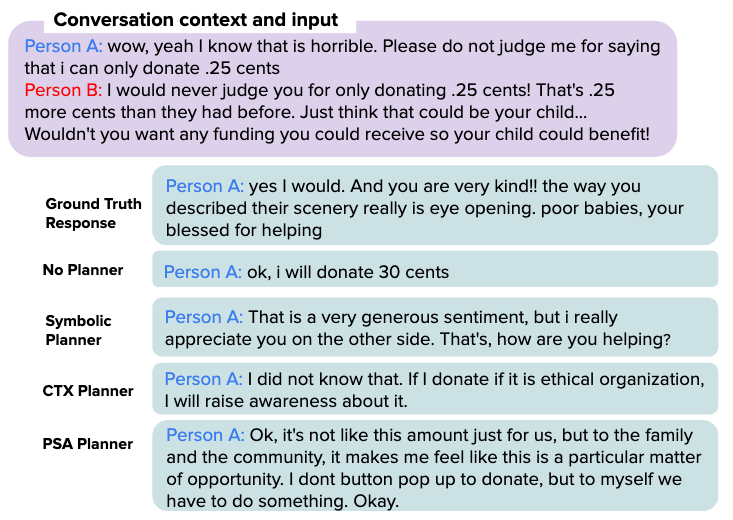}
    \caption{Third conversation example from the test set. In this example, we noticed that the PSA planner based model generates an incoherent response.}
    \label{fig:example_supple3}
\end{figure}

\begin{figure}[h]
    \centering
    \includegraphics[width=\columnwidth, height=7cm]{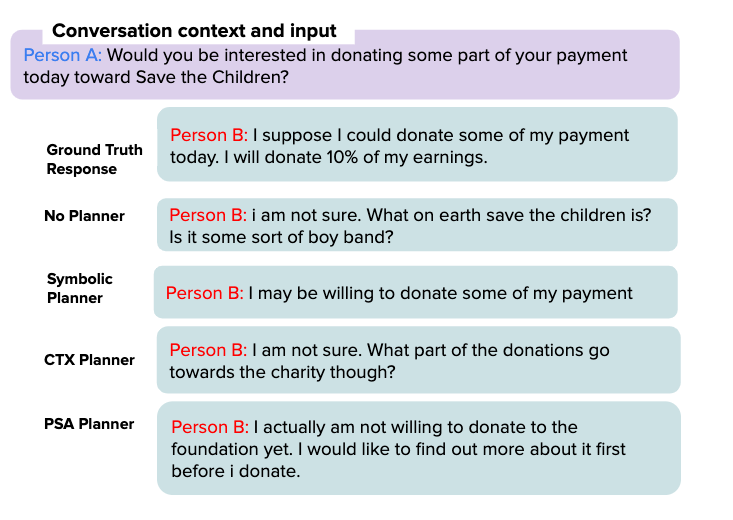}
    \caption{An additional conversation example with output from realizers with all four planner configurations. }
    \label{fig:example_supple4}
\end{figure}

\newpage


\begin{table*}[t]
\footnotesize

\begin{tabular}{@{}lll@{}}
\toprule
\multicolumn{1}{c}{\textbf{Issues Found}} & \multicolumn{1}{c}{\textbf{Input Utterance and Context}} & \multicolumn{1}{c}{\textbf{\begin{tabular}[c]{@{}c@{}}Generated Plan \\for Response/ \\ Generated Response\end{tabular}}} \\ \midrule
\multicolumn{3}{c}{Planner Phase} \\ \midrule
\midrule
\begin{tabular}[c]{@{}l@{}}\textit{Non-Informative} \\ \textit{Ask/Framing Target}\end{tabular} & \begin{tabular}[c]{@{}l@{}}I donate pretty regularly with my kids in school,\\ they have different charities they help out each month.\end{tabular} & PERFORM [look [ i ]] \\ \midrule

\begin{tabular}[c]{@{}l@{}}\textit{Non-Informative} \\ \textit{Ask/Framing Target}\end{tabular} & \begin{tabular}[c]{@{}l@{}}If you get time, go to Save the Children's website \\and check out more information. Maybe someone\\  you know would like to make a  donation \\or match your donation to help children in \\ other countries  affected by war and poverty.\end{tabular} & PERFORM [ support [them ]] \\ \midrule

\begin{tabular}[c]{@{}l@{}}\textit{Non-Informative} \\ \textit{Ask/Framing Target}\end{tabular} & \begin{tabular}[c]{@{}l@{}}Thats great to hear, empathy and sympathy goes \\ a long way and is lacking in society \\ so its great to hear another parent\\ trying to help out as much as they can.\end{tabular} & GIVE [ donate [to ]] \\ \midrule

\begin{tabular}[c]{@{}l@{}}\textit{Wrong Ask/Framing } \\ \& \textit{type and action}\end{tabular} & \begin{tabular}[c]{@{}l@{}}Have you felt inclined to donate to any charity\\ before? If so, what sort of charity?\end{tabular} & GIVE [give [ part ]] \\ \midrule \midrule

\begin{tabular}[c]{@{}l@{}}\textit{Wrong Ask/Framing } \\ \& \textit{type and action}\end{tabular} & \begin{tabular}[c]{@{}l@{}}I don't understand. They make money \\through online searches?\end{tabular} & GIVE [ give [ online ]] \\ \midrule \midrule \\

\multicolumn{3}{c}{Realization Phase} \\ \midrule \midrule

\begin{tabular}[c]{@{}l@{}} \textit{Ignored} \\
\textit{Plan}
\end{tabular} & \begin{tabular}[c]{@{}l@{}}Would you be interested in donating some of your \\ payment today toward Save the Children?\end{tabular} & \begin{tabular}[c]{@{}l@{}}Plan: GIVE[ donate[\$ 1 ]]\\ Response:  I am not sure. What\\ part of the donations go \\towards the charity though?\end{tabular} \\ \midrule

\begin{tabular}[c]{@{}l@{}} \textit{Ignored} \\
\textit{Plan}
\end{tabular} & \begin{tabular}[c]{@{}l@{}}Person A: Do you have a personal connection \\to the cause?\\ Person B: Yes. Family members and friends have gone \\through this. \end{tabular} & \begin{tabular}[c]{@{}l@{}}Plan: LOSE[ lost [my \\ grandmother ]]\\ Response:  I have a friend that \\ lost his parents at the \\ Mexico-US border. It broke my \\ heart\end{tabular} \\ \midrule

 \begin{tabular}[c]{@{}l@{}} \textit{Grammatical} \\
\textit{Inconsistency}
\end{tabular} & \begin{tabular}[c]{@{}l@{}}Person A: i want to assure you we \\are on an encrypted line. \end{tabular} & \begin{tabular}[c]{@{}l@{}}Plan: PERFORM[verify[that]]\\ Response: i can verify that,\\ please verify that.\end{tabular} \\ \midrule

 \begin{tabular}[c]{@{}l@{}} \textit{Grammatical} \\
\textit{Inconsistency}
\end{tabular} & \begin{tabular}[c]{@{}l@{}}Person A: could you verify your credit card \\ expiration date\\ Person B: what does it say on your end?
 \end{tabular} & \begin{tabular}[c]{@{}l@{}}Plan: PERFORM[ provide [that i \\need to verify it from you ]]\\ Response: i prefer to provide that\\ info,  so please provide that info.\end{tabular} \\ \bottomrule
\end{tabular}
\caption{Summary table of issues found through a qualitative analysis of the planning and realization phase outputs}
\label{tab:issues_supple}
\end{table*}

\end{document}